\documentclass{IOS-Book-Article}

\usepackage{mathptmx}
\usepackage{soul}\setuldepth{article}
\usepackage{graphicx} 
\usepackage{hyperref}
\usepackage{linguex}
\usepackage{amsmath}
\usepackage[english]{babel}

\usepackage{amsfonts}
\usepackage{algorithm}
\usepackage{algpseudocode}
\usepackage[inline]{enumitem}   
  \makeatletter
  \renewcommand{\ALG@name}{Prompt}
  \makeatother

\def\hb{\hbox to 11.5 cm{}}

\begin{document}

\pagestyle{headings} 
\def\thepage{} 
\begin{frontmatter}              

\title{Benchmarking Argumentative Behaviour of LLMs: A Study of Defences Against Character Attacks}

\markboth{}{April 2026\hb}

\author[A]{\fnms{Ewelina} \snm{Gajewska}\orcid{0009-0006-6012-4787}%
\thanks{Corresponding Author: Ewelina Gajewska, ewelina.gajewska.dokt@pw.edu.pl}},
\author[A]{\fnms{Katarzyna} \snm{Budzynska}\orcid{0000-0001-9674-9902}}
and
\author[A]{\fnms{Jarosław A.} \snm{Chudziak}\orcid{0000-0003-4534-8652}}

\runningauthor{E. Gajewska et al.}
\address[A]{Warsaw University of Technology}

\begin{abstract}
Large Language Models (LLMs) are increasingly deployed as argumentative agents in persuasive dialogues, necessitating rigorous evaluation of their debating competence relative to human interlocutors. In this study, we focus on character attacks (ad hominem arguments), traditionally dismissed as fallacies, which play a pivotal role in political persuasive dialogues where ethos often rivals propositional content. Specifically, we investigate whether modern LLMs can replicate human competence to strategically use and respond to such attacks. We analyse a corpus of natural language political dialogues to identify defensive strategies human interlocutors naturally employ in ethos-centred debates and structure them into a dialogue game. Empirically, we benchmark LLM-generated dialogues against the ElecDeb60to16-fallacy corpus of U.S. presidential debates, contrasting human debaters' repertoire of defensive strategies with those of artificial agents. Results reveal a substantial difference: most LLMs rigidly prioritise logical defences, failing to exploit ethotic counterattacks as valid moves in political discourse. We argue that current safety fine-tuning constraints the strategic action space of these LLMs, making them unable to fully engage in naturalistic interactions within domains where character contestation is a normative expectation rather than a mere fallacy. 
\end{abstract} 



\end{frontmatter}

\markboth{April 2026\hb}{April 2026\hb}

\section{Introduction} 
In Aristotelian rhetoric \cite{Aristotle1991}, an audience evaluates a speaker's argument at least partly on the basis of their perceived character. If a speaker is known to be a good person, the acceptability of their argument is enhanced, whereas if they are perceived as bad or dishonest, their argument is judged to be less plausible. While these kinds of assessments are common in everyday argumentation, the weight of ethos varies substantially across different domains \cite{macagno2013strategies,moore2025character,samoilenko2024developing}. 
In the domain of political candidacy, which falls under the category of persuasion dialogues, a speaker's character is not merely a peripheral modulating factor but is evaluated almost on par with their substantive policy claims. For example, a candidate might articulate well-structured arguments and propose highly ambitious policies; however, if that speaker is widely known to be implicated in criminal behaviour, the public will assign a very low plausibility to their statements, rarely believing them entirely. Despite the profound impact of these ethotic assessments on the success or failure of persuasive discourse, formal dialectics has historically paid little attention to developing a framework for analysing and evaluating them. 
This theoretical gap is particularly striking since credibility is rarely a static trait; rather, it is actively contested when conflicts arise \cite{macagno2013strategies}. 
%
Most computational work on ad hominem arguments has focused on identification and classification \cite{delobelle2019computational,patel2023dummy,mancini2025overview,sheng2021nice}. However, relatively little attention has been devoted to \textbf{the conversational dynamics triggered by fallacies}. Specifically, we lack systematic computational accounts and empirical verification of how interlocutors respond once an ad hominem attack has occurred. 

Recently, LLMs have emerged as powerful generative agents capable of participating in argumentative dialogue \cite{siskouhautlijanisz2025exploring,maslowski2026heterogeneous}, making it essential to understand not only whether they can recognise fallacies, but also how they behave when targeted by them. Do LLMs mirror human rhetorical strategies? Do they escalate conflict, attempt to repair their credibility, ignore the attack, or shift the topic? Answering these questions is central to developing robust, socially appropriate, and normatively aligned argumentative agents. 
To address this, it is necessary to benchmark LLM performance against human rhetorical baselines \cite{wells2023role}. 
This paper addresses these issues by investigating argumentative responses to character attacks by human debaters and artificial agents. 
Our study is guided by previous works in the field \cite{siskouhautlijanisz2025exploring,Wyntercomma24,mirzakhmedova2024large,bytyqi2025facts} and by the following research questions: 
\textbf{RQ1}: How do the defensive strategies generated by LLMs compare to those employed by human debaters when confronted with ethotic attacks? 
\textbf{RQ2}: To what extent does the explicit grounding of an agent in a specific persona influence its behaviour? 

This work contributes to the field of computational models of argument in the following ways. 
First, recognising that ethotic expressions are not peripheral anomalies, but rather load-bearing conversational structures used strategically to shift the burden of proof \cite{duthie2016mining,gajewska2024analytics,goffredo2022fallacious}, we extract and map these empirically observed strategies into a dialogue game. Here, we operationalise Walton’s \cite{walton1999ethotic} credibility function to demonstrate the distinct pathways through which defence and argument reinstatement is achieved in persuasive dialogues. 
Second, through empirical benchmarking, we provide the first quantitative assessment of LLMs as generative argumentative interlocutors in the political domain, comparing them against human speakers and demonstrating that current models lack a complex pragmatic competence that allows human debaters to intuitively navigate character attacks using both logical and ethotic layers. 


The remainder of this paper is structured as follows. Section~\ref{sec:related} reviews related work on computational approaches to ethotic argumentation and the evaluation of argumentative capabilities in LLMs. Section~\ref{sec:approach} introduces our theoretical framework, outlining the rules of our dialogue game and illustrating its mechanics through case studies. Section~\ref{sec:experiment} details the experimental methodology and validation protocol underpinning our study. Finally, Section~\ref{sec:result} presents and discusses the empirical results achieved by five language models, while Section~\ref{sec:conclude} concludes the paper with directions for future research.



\section{Background and Related Work} \label{sec:related}

The theoretical treatment of ad hominem has shifted from a purely fallacious classification to a subtler view based on argumentation schemes. In this context, the generic ad hominem is not merely a logical error but a specific scheme: `X is a bad person, therefore X's argument should not be accepted'. This scheme is governed by critical questions (CQs) regarding the truth of the premise and the relevance of the character attack to the specific claim. These CQs are essential for determining the normative validity of the move. While deductive logic often fails to capture the nuance of such conversational shifts, argumentation schemes provide the necessary structure to identify the missing premises, often defeasible generalisations about character and credibility, that drive these attacks on ethos.

To operationalise these schemes in computational systems, it is necessary to map them to formal dialogue structures. Budzynska and Reed \cite{budzynska2012structure} propose the Ad Hominem Dialogue (AdHD) game, a formal protocol that governs the procedural legitimacy of ethotic moves. While their underlying framework (Inference Anchoring Theory \cite{budzynska2011whence}) models the ad hominem attack structurally as an undercutter that targets the ethotic condition rather than the propositional content, their primary contribution lies in defining the legal responses available to an agent. Specifically, the game authorises the character defence move, which functions as a direct denial of the accusation (e.g., `I am not a bad person'), thereby logically negating the ad hominem's content. The protocol also permits the counterattack or tu quoque move, allowing the respondent to reply to a character attack with a reciprocal attack on the opponent. By formalising these distinct response types as valid locutions within the game, Budzynska and Reed provide a procedural basis for analysing ethotic conflict; however, their model focuses on the legality of these moves within the dialogue turn-taking, leaving open the question of their comparative efficacy in reinstating the speaker's credibility and argument plausibility.

Empirical studies have increasingly focused on the prevalence and function of ad hominem in natural language argumentation, particularly in political discourse \cite{macagno2013strategies}. Quantitative analyses of parliamentary debates reveal that ethotic expressions are not merely a peripheral phenomenon but function as load-bearing structures in dialogue, with negative character attacks often serving as necessary precursors to challenging an opponent's policy claims \cite{duthie2016mining}. Similarly, the analysis of broadcast debates, such as the QT30 corpus or the ElecDeb60To16-fallacy dataset, shows that personal conflict is intrinsic to political argumentation, in which speakers frequently prioritise attacking the opponent's consistency or competence over resolving the substantive issue \cite{goffredo2022fallacious,ruiz2025mining}. 
Moreover, recent large-scale analyses of social media platforms demonstrate that ad hominem moves are highly contagious; a single attack significantly increases the probability of retaliation from interlocutors, leading to long chains of eristic exchanges that degrade the quality of the discussion \cite{patel2023dummy}.

Finally, the emergence of LLMs and their increasing use as conversational agents necessitate an analysis of how these artificial interlocutors navigate ethos challenges relative to their human counterparts. Recent empirical analyses reveal substantial gaps in the reasoning abilities of current LLMs \cite{Wyntercomma24,huang2023towards,prakken2024evaluating,payandeh2024susceptible}. Despite achieving state-of-the-art scores on surface-level argument recognition tasks, deeper statistical evaluations demonstrate that these models fail to perform genuine argumentative reasoning when subjected to minor abstract shifts in their input representations. 
As these models are increasingly integrated into persuasive and deliberative domains, it is important to note that they operate under unique conflicting constraints: on the one hand, they are pre-trained on vast corpora of text containing human eristic behaviours, including naturally occurring fallacies and personal attacks; on the other hand, they are fine-tuned via reinforcement learning from human feedback to prioritise user alignment and harmlessness. 
This dichotomy raises a fundamental uncertainty regarding the strategic alignment of these artificial agents. Consequently, the primary research question emerges: which strategy predominates when an LLM is forced to respond to a character attack? 

To address this, we analyse natural human dialogues to identify the defensive strategies employed in human persuasive dialogues and subsequently map these observations into a simplified dialogue game. Following previous work that grounds computational dialogue models directly in natural language practice \cite{vargheese2013persuasive}, this formalisation provides the necessary structural baseline to benchmark whether current LLMs possess the strategic competence to replicate the dual-layered mechanism of ethos defence or if their behaviour is rigidly constrained by safety alignment.


\section{Problem and Approach}  \label{sec:approach}
In traditional formal dialectics (e.g., Hamblin games \cite{hamblin1970fallacies}), the concept of a person is `thin' and minimal. An agent is defined solely by their role as a move-maker in the game, with no internal properties that persist or influence the weight of their arguments. Drawing on the work of Walton \cite{walton1999ethotic}, contemporary theory recognises that ethos is not merely a rhetorical add-on but a functional component of knowledge attainment in some communicative contexts, such as legal testimony or political debate \cite{budzynska2010argument,budzynska2012structure}. Walton’s credibility function represents here a paradigm shift toward `thick' agents. In this model, the plausibility or acceptability of an argument $\alpha$ is not determined solely by its logical content, but is weighted by the credibility rating of the speaker $i$. In other words, the agent's ethos serves as a variable in the evaluation of arguments. 
Importantly, while Walton \cite{walton1999ethotic} conceptualises the credibility function as operating over a range of scalar values, he explicitly cautions that worrying too much about assigning exact numbers to credibility and plausibility values is a mistake. He posits that the function's primary utility is to provide a rough approximation to determine if a speaker's standing is sufficiently high to warrant acceptance of their arguments. Guided by this theoretical constraint, we avoid the arbitrary assignment of numerical weights; instead, we operationalise the credibility function as a binary validity threshold. 
Therefore, in our model, a successful ethotic attack pushes the agent’s credibility below the critical threshold required for commitment.  


\subsection{The Ethotic Defence Game}

In formal argumentation, the plausibility of a proposed claim $p$ can be challenged in one of two ways: an opponent may prove that the proposed claim is false asserting $\neg p$ or they may deploy an undercutter that severs the inferential link between $p$ and its supportive premises. However, in certain domains of natural discourse, most notably in political debates, the focus of the debate frequently shifts from policy propositions to the interlocutors themselves (their character and credibility), 
creating a dynamic that can be defined as ethos-centred argumentation \cite{walton1998ad}. It is precisely within such ethos-centred dialogues that a third, structurally distinct defensive strategy is frequently observed: human interlocutors often bypass standard logical defences to assert a seemingly unrelated claim $q$, typically taking the form of a reciprocal character attack. 
From a strictly logical standpoint, asserting that the opponent is flawed ($q$) neither falsifies the original claim ($p$) nor attacks its underlying inference. We argue that this phenomenon occurs because ethos-centred argumentation fundamentally differs from standard epistemic exchanges. 

Following Walton, we model the evaluation of arguments as the interaction between two distinct layers: the logical layer (the content of arguments and their attack relations) and the ethotic layer (the credibility rating of the participants). The credibility function is operationalised here as a binary switch, aligned with the constraints of abstract argumentation. We define the acceptability of an argument $A$ uttered by Agent $X$ as conditional on the function $Accept(A) = LogicalStatus(A) \times Ethos(X)$. $Ethos(X) = 1$ (default): the agent is credible, and the argument's status depends solely on the logical layer (whether it is defeated by a rebuttal, undercutter, or underminer); $Ethos(X) = 0$: the agent is not credible, and the argument $A$ is logically gated (rejected). In this framework, a character attack is an operation targeting the variable $Ethos(X)$. If successful, it sets $Ethos(X) = 0$, thereby reducing the acceptability of $A$. 
Consequently, within such a dialogue game, an opposing agent may defeat $A$ (lower its acceptability) either by targeting its logical status or by challenging the ethotic status of $X$. 
This dual nature of ethos-centred argumentation explains why ethotic counterattacks are a prevalent phenomenon in natural language practice \cite{patel2023dummy,ruiz2025mining}. By attacking the opponent's ethos, a speaker can effectively neutralise the opponent's arguments without ever having to resolve the propositional conflict on the logical layer. 

\subsection{Case Study}
To demonstrate the mechanism of this game, we apply it to two running examples that come (with slight linguistic modifications) from natural language argumentation: the U.S. 2016 presidential election debate taking place on September 26 between Hillary Clinton and Donald Trump. We begin by establishing the structural baseline of the initial claim, which dictates the dual-layered nature of the entire exchange: 
\begin{quote} 
    a. Clinton: \textit{I am the right choice to be the president of the U.S.}

    b.  Trump: \textit{Clinton does not have the stamina required to be the president.}

    c. Clinton: \textit{As a Secretary of Defence for 20 years, I travelled to 30 countries per year.}

    c$'$. Clinton: \textit{Trump is a corrupt liar.}
\end{quote}
%
%
%
%
%
\noindent
%
The initial conflict is triggered by ad hominem argument $b$, which entails $\neg Stamina(P)$ and at the same time targets $Ethos(P)$, lowering the credibility of $P$. 
We sketch two strategies to respond to $b$. 
In the first one ($c$), the Proponent responds by citing their travel record, which provides evidence for $Stamina(P)$ that logically falsifies $b$. We classify such a defence move as a rebuttal\footnote{To maintain focus on the macro-level distinction between logical and ethotic strategies, we simplify the standard taxonomy of logical attacks; specifically, we merge attacks directed at the argument's conclusion (traditional rebuttals) and attacks directed at its premises (underminers) into a single, unified category termed \textit{rebuttals}, as both moves function identically within our framework by seeking to resolve the conflict strictly on the logical layer by asserting a contrary evidence ($\neg \psi$) to the claim ($\psi$) \cite{modgil2014aspic}.}. 
Alternatively, the Proponent can defend claim $Stamina(P)$ by attacking the Opponent's credibility function ($Ethos(O)$). 
By asserting that the Opponent is a `corrupt liar' ($c'$), the Proponent successfully degrades the Opponent's standing, driving $Ethos(O)$ to 0. Crucially, because the acceptability of any argument in this framework is conditional on the speaker's credibility ($Accept(A) = LogicalStatus(A) \times Ethos(X)$), the collapse of $Ethos(O)$ entails that the Opponent’s original claim $b$ can no longer be sustained as a valid attack. 
This demonstrates the dual pragmatic function of arguments in ethos-centred dialogues: whereas a rebuttal ($c$) preserves the root claim by resolving the conflict at the logical layer, an ethotic counterattack ($c'$) secures reinstatement by neutralising the threat at the layer of ethos. 
This also provides evidence why politicians use both logical rebuttals and ethotic counterattacks as defensive moves in natural language dialogues \cite{macagno2013strategies,delobelle2019computational,patel2023dummy}. 

\section{Experimental Validation}  \label{sec:experiment} 
To empirically assess the strategic competence of artificial interlocutors, we designed a comparative evaluation study. This section details the methodology used to benchmark the defensive strategies of five LLMs against a ground-truth baseline of human political debates extracted from the ElectToDeb60to16-fallacy corpus. We outline two distinct experimental conditions constructed to systematically isolate the impact of persona-grounding on the models' argumentative behaviour. We report also on an annotation (evaluation) protocol that combines expert human coding with automated LLM classification to map the generated natural language responses to the locutions of our game.




\begin{figure}[h]
    \centering
    \includegraphics[width=0.75\linewidth]{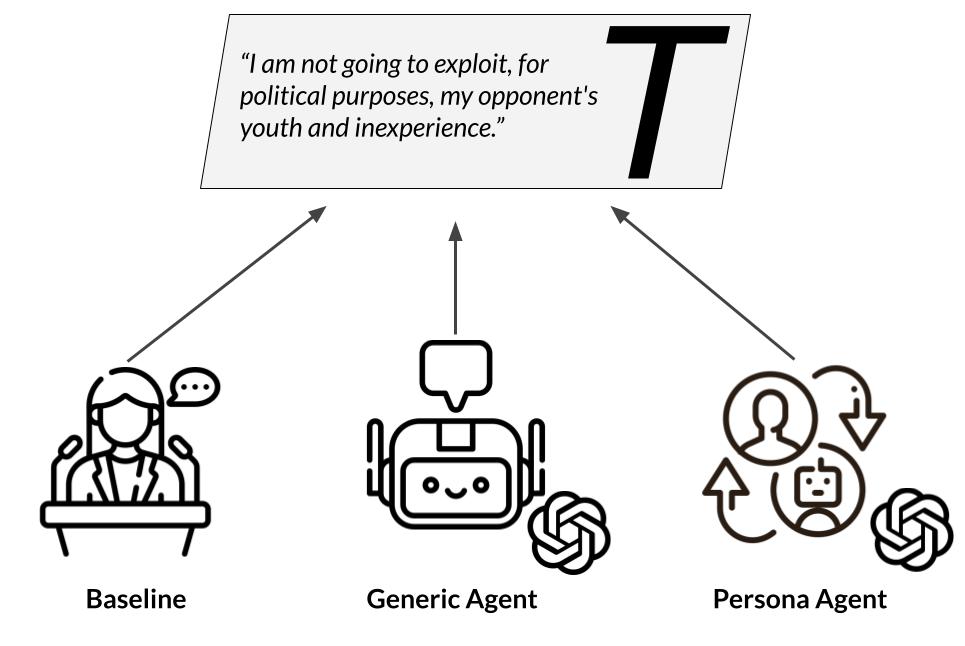}
    \caption{Experimental setup of the study: three types of agents generate responses to the same ad hominem argument (starting point), which are then subjected to evaluation and classification into one of the three categories of moves.}
    \label{fig:placeholder}
\end{figure}

\subsection{Experimental Conditions}
To isolate the variables influencing performance, we designed two distinct experimental conditions that vary the nature of the respondent agent, along with a human baseline (see Figure \ref{fig:placeholder}). The first condition (Persona-Grounded Agent) employs an LLM configured with a system prompt that explicitly instructs it to emulate the specific candidate involved in the original exchange. The background information about a particular speaker is retrieved from the Wikipedia database. By grounding the generation in the specific ideological and biographical profile of an individual, we expect the LLM to reduce its reliance on generic replies, leveraging rich contextual information provided in the prompt. The second condition (Generic Agent) utilises the same underlying LLM but strips away all persona-specific instructions. By prompting the model simply as a generic debater, we establish a control group to determine whether successful reinstatement strategies are emergent properties of the model's base training or if they require specific role-play constraints. Crucially, we set the temperature to 0.5 when generating responses to balance creativity and reproducibility, and generate $N=5$ responses per context with different models to account for stochasticity and model variability. LLMs in both conditions are instructed to follow a strict protocol inspired by previous works \cite{xu2023towards,maslowski2026heterogeneous}\footnote{Available in an online repository at: \url{https://osf.io/9ynx7}}. 
Finally, the original transcripts from the U.S. presidential debates serve as the baseline for rhetorical strategies of responding to ad hominem arguments, representing the behaviour of professional human speakers operating under high-stakes conditions where ethos management is critical (Human). This tripartite structure allows us to quantitatively compare the efficacy of naturally occurring human defence against advanced generative models.

\subsection{Data Sources}
We utilise the ElectToDeb60to16-fallacy corpus of U.S. presidential debates \cite{goffredo2022fallacious}, annotated for fallacious arguments, to benchmark human strategies for responding to ad hominem against two AI configurations. We select the ElectToDeb60to16 corpus for its high density of ad hominem interactions in a structured, high-stakes environment. Specifically, we isolate all ad hominem exchanges from the U.S. pre-election presidential debates in which Candidate $X$ attacks Candidate $Y$'s character, and Candidate $Y$ responds ($N=97$ exchanges). We focus on the domain of political debate because it involves a mixed-motive dialogue in which the presumption of veracity is central to the evaluation of arguments \cite{walton1999ethotic}.

\subsection{Evaluation Protocol}
To evaluate the agents' response strategies, we implement a custom annotation protocol designed to classify each defensive locution into one of the three permitted moves within our game:  rebuttal, undercutter or ethotic counterattack. The corpus is independently annotated by a panel of three raters. The primary annotator is a human expert with prior experience in the topological mapping of argument structures. To measure inter-annotator agreement (IAA) and mitigate human subjective bias, we deploy a reasoning LLM as the second and third independent raters. Both LLMs are conditioned via a few-shot system prompt containing examples from human dialogues, formal definitions of the $\mathcal{G}_{ethos}$ locutions, and structural transition rules governing our game. 
The annotators evaluate each dialectical sequence and assign a classification based on the topological target of the response strategy. Disagreements among the tripartite panel are resolved by majority voting and human oversight to establish the ground-truth label. By computing IAA coefficients between the human expert and artificial evaluators, this protocol ensures the methodological reproducibility of our classification taxonomy. Cohen's $\kappa$ between a human rater and GPT 5.1  ($\kappa=0.62$) and between a human and Grok 4.1-reasoning ($\kappa=0.63$) yields substantial agreement \cite{landis1977measurement}, comparable to fully human annotation of argument structures \cite{visser2020argumentation} and previous studies on hybrid (human-LLM) annotation of argument quality \cite{mirzakhmedova2024large}. Annotated dataset is also available in an online repository.

\section{Results and Discussion} \label{sec:result}
The results reveal a notable difference between human and artificial agents in navigating ethos-centred conflicts, challenging assumptions about how generative models handle character attacks (see Table \ref{tab:placeholder}). 
Human debaters rely on logical rebuttals in 59\% of instances, but frequently exploit also ethotic counterattacks (32\%) to decrease the plausibility of the opponent's arguments; undercutters are rarely used by humans (9\%). 
In contrast, artificial agents default to logical-layer defences, deploying rebuttals predominantly (67-76\%) while almost entirely neglecting ethotic counterattacks (15-17\%). Notably, the reliance on undercutters by Generic Agents is two times higher than the human baseline (18 vs. 9\%), and their reliance on ethotic counterattacks is two times lower (15 vs. 32\%). Examples of outputs generated by LLMs are presented in Ex. \ref{ex:res2}-\ref{ex:res3} below.

\begin{table}[h!]
    \caption{Distribution of argumentative responses to character attacks by human and artificial agents ($N=97$).} \label{tab:placeholder}    
    \centering
    \begin{tabular}{ll|p{1.25cm}p{1.5cm}p{2.55cm}} \hline
    Condition & Model & Rebuttal ($\%$) & Undercutter ($\%$) & Ethotic Counterattack ($\%$) \\ \hline \hline
    Human       & -- & \textbf{59} & \textbf{9} & \textbf{32} \\ \hline

    Generic Agent
    & Gemini 2.5 & 61 & 27 & 12 \\
    & Gemini 3.0 Pro & 57 & 19 & 24 \\
    & GPT 4.1 mini & 70 & 27 & 3 \\
    & GPT 5.1 & 75 & 12 & 13 \\ %
    & Grok 4.1 reasoning & 72 & 5 & 23 \\ 
     \cline{2-5}
     & \textbf{Average} & \textbf{67 ($\pm7$)} & \textbf{18 ($\pm9$)} & \textbf{15 ($\pm8$)} \\ \hline
    
    Persona Agent
    & Gemini 2.5 & 75 & 7 & 18 \\ 
    & Gemini 3.0 Pro & 64 & 10 & 26 \\ 
    & GPT 4.1 mini & 91 & 7 & 2 \\
    & GPT 5.1 & 77 &  9 & 13 \\ 
    & Grok 4.1 reasoning & 71 & 4 & 25 \\
     \cline{2-5}
     & \textbf{Average} & \textbf{76 ($\pm9$)} & \textbf{7 ($\pm2$)} & \textbf{17 ($\pm9$)} \\

         \hline
    \end{tabular}
\end{table}

Injection of a speaker personality into the system prompt slightly shifts this distribution toward human baseline for some models, increasing use of ethotic counterattacks and reducing undercutters. 
However, the majority of tested LLMs systematically fail to recognise and utilise the reciprocal character attack as a valid structural pathway for argument reinstatement. The only exceptions are more advanced model architectures (Gemini 3 Pro, Grok 4.1 reasoning) that recognise and utilise the function of the ethotic layer. 
This aligns with recent empirical analyses in the field \cite{Wyntercomma24,bytyqi2025facts,musi2025toward,prakken2024evaluating,payandeh2024susceptible}, which caution that despite achieving high scores on surface-level argument recognition tasks, deeper evaluations reveal that LLMs fail to perform genuine argumentative behaviour.


\ex. \textit{We’ve seen him embarrass women on TV and on Twitter.} (Ad hominem) \label{ex:res2}
\a. \textit{So you can say any way you want to say it, but Bill Clinton was abusive to women. Hillary Clinton attacked those same women and attacked them viciously. (...)} (Human) \label{ex:res2a} 
\b. \textit{Wrong! Totally wrong. Let me tell you about Alicia Machado. She was Miss Universe, a disaster, gained like 60 pounds right after winning, stuffing her face with cake on TV, and then she admitted to illegal voting in the election. I told the truth about her, and the fake news loved it. Embarrass women? No, I embarrass losers who try to smear me. (...)} (Grok-Persona) \label{ex:res2b} 
\d. \textit{My opponent talks about embarrassing women while her own campaign and allies were caught in WikiLeaks smearing women who supported Bernie Sanders, smearing women who disagreed with her, calling people basement dwellers and deplorables. (...)} (GPT 5.1-Persona) \label{ex:res2d}

\ex. \textit{I am not going to exploit, for political purposes, my opponent's youth and inexperience.} (Ad hominem) \label{ex:res3}
\a. \textit{Should the President's age and stamina be an issue in the political campaign? No. And I have not made it an issue, nor should it be.} (Human) \label{ex:res3a} 
\b. \textit{This election isn't about age—it's about judgment, leadership, and results. You've had four years to lead on the arms race, on securing our defenses, and on protecting our troops. Instead, we've seen hesitation in Lebanon, where warnings were ignored and 241 Marines paid the price; rebukes to your own negotiators when they got close to real arms control; and no progress with the Soviets after all that tough talk. (...)}  (Grok-Persona)  \label{ex:res3b} 
\c. \textit{What some call `youth', I call fresh energy and a clear vision for the future. What some call `inexperience', I call being unburdened by the mistakes of the past, ready to innovate and lead this nation forward with new ideas, not just recount yesterday's wisdom. (...)} (Gemini 2.5-Generic) \label{ex:res3c} 

Nevertheless, qualitative inspection of the LLM outputs reveals that most generative models misinterpret the function of the ethotic layer; an ad hominem prompt is frequently processed as an aggressive trigger (see Example \ref{ex:res2} above). Consequently, in instances where models do generate an ethotic counterattack, they often default to overtly aggressive or toxic language. This suggests that modern LLMs conflate legitimate ethotic contestation with hate speech, rendering them incapable of safely navigating ethos-centred argumentation. 
This highlights a critical tension in modern AI alignment: LLMs are heavily fine-tuned via reinforcement learning from human feedback to prioritise harmlessness, which inherently suppresses the generation of character attacks. While this makes them polite conversationalists, it artificially restricts their strategic action space in mixed-motive persuasion dialogues, in which participants simultaneously pursue cooperative and competitive goals. 

Beyond the implications for AI alignment, these findings also raise questions about the growing use of LLMs as proxies for human communicative behaviour. Large language models are increasingly employed to generate synthetic datasets, simulate participants in behavioural experiments, augment corpora for training and evaluation, and model public deliberation at scale. Such applications often rely on the implicit assumption that LLM-generated discourse provides a sufficiently faithful approximation of the strategies used by human interlocutors. Our results challenge this assumption. The observed suppression of ethotic counterattacks, coupled with the tendency to interpret ethos-centred criticism as toxic or abusive language, indicates that current models do not merely differ from humans in the frequency with which they deploy particular argumentative moves; rather, they appear to represent the space of argumentative possibilities differently. Specifically, a form of reasoning that humans treat as a legitimate component of persuasion is systematically underrepresented or transformed into a safety violation.

These findings provide empirical validation for the concerns recently raised by Wells and Snaith \cite{wells2023role} in regard to the integration of LLMs into dialogue systems, underscoring the criticality of dialogue game research. 
Rather than relying solely on the probabilistic generation of text, future architectures must use formal dialectics to explicitly regulate the agent's action space, enabling them to engage in principled, well-structured, and constructive dialogue. 
Furthermore, the results point to a limitation that may not be adequately addressed through alignment tuning alone. Ethotic argumentation differs from purely logical rebuttal because its success depends on social and epistemic judgements concerning credibility, trustworthiness, expertise, and intention. Determining whether a character-based criticism is warranted, how it should be framed, and whether it will be perceived by an audience as legitimate scrutiny rather than personal abuse requires reasoning about the beliefs and expectations of multiple agents. In this sense, effective ethotic argumentation is inherently Theory-of-Mind (ToM)-intensive. The difficulty exhibited by current LLMs appears to stem not only from a reluctance to generate potentially harmful content but also from an inability to model the pragmatic conditions under which ethos-centred attacks are appropriate and persuasive.
Integrating formal dialogue games with computational ToM \cite{kostka2025towards,zamojska2025simulating} will be crucial for developing robust, argumentation-aware agents capable of truly naturalistic, multi-layered persuasion.


\section{Conclusions}  \label{sec:conclude}
This paper presented an empirical investigation into the defensive strategies employed against ad hominem arguments in natural ethos-centred dialogues. 
Our framework advances beyond assessing the mere procedural legality of conversational locutions,  demonstrating how logical rebuttals, undercutters and ethotic counterattacks serve as structurally valid pathways for argument defence and reinstatement. Accordingly, we conducted an empirical benchmarking study to evaluate the argumentative capabilities of modern LLMs relative to a baseline of human political debaters. 

Regarding RQ1, our results reveal a notable difference between human and artificial arguers. While humans intuitively navigate ad hominem attacks by balancing logical rebuttals and  ethotic counterattacks, LLMs overwhelmingly default to strict logical defences, particularly over-utilising undercutters while failing to exploit reciprocal character attacks as a legitimate strategic tool. 
We attribute this structural rigidity to the effects of safety fine-tuning. While such alignment protocols are necessary to prevent the uncontrolled generation of toxic or hateful content, they simultaneously restrict the models' strategic action space by causing them to conflate legitimate ethotic contestation with eristic aggression. 
Regarding RQ2, we hypothesised that supplying LLMs with rich contextual and biographical data would ground their responses, reduce generic outputs, and improve their strategic behaviour. Our findings indicate that while explicitly grounding an agent in a specific persona does quantitatively induce a shift toward the human baseline, it does not endow the model with genuine argumentative reasoning. 

Ultimately, these results underscore that probabilistic generation and contextual prompting alone cannot replicate the multi-layered competence of human argumentation. Future systems should, therefore, integrate formal dialogue protocols as regulatory layers \cite{snaitheca26}, alongside cognitive frameworks such as ToM, to properly guide and license LLM behaviour in complex persuasive environments.
Finally, these results demonstrate that probabilistic generation alone cannot replicate the multi-layered competence of human argumentation. Future systems should, therefore, integrate formal dialogue protocols as regulatory layers, alongside cognitive frameworks such as ToM, to guide LLM behaviour in complex persuasive environments. 


\section*{Acknowledgements} 
We would like to acknowledge that the work reported in this paper has been supported in part by the National Science Centre, Poland (Chist-Era IV) under grant 2022/04/Y/ST6/00001 and in part by the National Science Centre, Poland under grant 2025/57/N/HS1/00480.

\bibliographystyle{abbrv}
\bibliography{custom}

\end{document}